# The Application of Fuzzy Logic to Collocation Extraction

[1]Raj Kishor Bisht[*],  [2]H.S.Dhami

[1]*Department of Applied Sciences, Amrapali Institute of Technology and Sciences, Haldwani, Uttarakhand (India)*
[2]*Department of Mathematics, University of Kumaun .S.S.J.Campus Almora, Uttrakhand (India)*

**Abstract**

Collocations are important for many tasks of Natural language processing such as information retrieval, machine translation, computational lexicography etc. So far many statistical methods have been used for collocation extraction. Almost all the methods form a classical crisp set of collocation. We propose a fuzzy logic approach of collocation extraction to form a fuzzy set of collocations in which each word combination has a certain grade of membership for being collocation. Fuzzy logic provides an easy way to express natural language into fuzzy logic rules. Two existing methods; Mutual information and t-test have been utilized for the input of the fuzzy inference system. The resulting membership function could be easily seen and demonstrated. To show the utility of the fuzzy logic some word pairs have been examined as an example. The working data has been based on a corpus of about one million words contained in different novels constituting project Gutenberg available on www.gutenberg.org. The proposed method has all the advantages of the two methods, while overcoming their drawbacks. Hence it provides a better result than the two methods.

## 1. Introduction

'Collocations' are a class of word groups which lie between idioms and free word combinations. However, it is typical to draw a line between a phrase and a collocation. Idioms and phrases may be defined as expressions in the language that is peculiar to itself either grammatically or especially in having a meaning that cannot be derived from the sum of the meanings of its elements. It becomes well nigh impossible to guess the meaning of an idiom from the words it contains (e.g. At the eleventh hour). And, moreover, the meanings that idioms have are often stronger than the meanings of non-idiomatic phrases. For instance, 'look daggers at someone' is more emphatic than 'look angrily at someone', although both of them have the same meaning [5]. On the other hand, in a free word combination, a word can be replaced by another word without seriously modifying the overall meaning of the composite unit so that one can not easily predict it from the remaining ones. For example 'end of the day', can not be predicted from 'end of the lecture' if we replace 'day' by 'lecture'. According to Kathleen R. McKeown and Dragomir R. Radev [8] 'collocations are arbitrary, language specific, recurrent in context and common in technical language'.

*Corresponding author.
 E- mail address:  bisht_rk@rediffmail.com

Collocations are utilized for many natural language applications such as, machine translation, computational lexicography, information retrieval, natural language generation etc. Collocation translation improves the quality of machine translation. Automatic identification of important collocations to be listed in a dictionary is the task of computational lexicography. Adequate knowledge of collocations can improve the performance of information retrieval system.

Statistical methods have shown a remarkable presence in collocation extraction. Frequency measure was used by Choueka et al [2] to identify a particular type of collocations. Church and Hanks [3] used mutual information to extract word pairs that tend to co-occur within a fixed size window (normally 5 words), in which extracted words may not be directly related. Smadja [11] extracted collocations through a multi-stage-process taking the relative positions of co-occurring words into account. Church and Gale [4] used the $\chi^2$- test for the identification of translation pairs in aligned corpora. Collocations extraction and their use in finding word similarity was suggested by Dekang Lin [9]. The use of t-test to find words whose co-occurrence patterns best distinguish between two words was suggested by Church and Hanks [3]. Dunning [6] applied likelihood ratio test to collocation discovery. Manning et al [10] may be referred for the detail discussion of collocation extraction is given in.

Decision of word combination for being collocation is a vague measurement. In the first paper on fuzzy decision making Bellman and Zadeh [1] suggest a fuzzy model of decision making in which relevant goals and constraints are expressed in terms of fuzzy sets and a decision is determined by an appropriate aggregation of these fuzzy sets. Fuzzy logic provides an easy way to check the possibility whether a word combination can be considered as collocation or not. Fuzzy logic allows the formation of a logic based model by utilizing the reasoning behind the existing methods. The resulting model has the simplicity of the logic based model and performs better than the existing statistical models. One may refer Klir et al [7] for detail discussion of fuzzy sets , fuzzy logic and fuzzy decision making.

## 2. Motivation

The existing statistical methods of collocation extraction based on the independence property of the two random variables, i.e., if two random variables $x$ and $y$ are independent, then $P(x.y) = P(x).P(y)$. Dependence of two words provides a chance to look for collocation. Almost all the techniques of collocation extraction look at whether the probability of seeing a combination differs significantly from what we would expect from their component words and reject those word combinations that do not have significant difference. We have taken two methods into consideration, Mutual information and t-test. High mutual information shows the presence of collocation while t-score provides a criterion for the selection of collocation which depends on the level of significance of the hypothesis. Both of the methods provide binary decision criterion {relevant, nor relevant or collocation, non-collocation}, which does not provide a fair chance to every word combination to be considered for collocation.

Fuzzy logic is a logical system which is very easy to understand. Word combinations getting scored by mutual information and t-score are evaluated by the rules of fuzzy inference system (FIS). The rules that are used to determine relevance of a word combination, come from the reasoning behind the existing techniques (e.g., if the mutual information score is high, then the word combination is highly relevant to be considered for collocation; if the t-score is significant, then word combination is relevant to be considered for collocation, etc.) Fuzzy logic expresses relevance as degrees of memberships (e.g., word combination could have a relevance measure with the following degrees of membership: 0.9 highly relevant and 0.5 reasonably relevant and 0.1 not relevant).

## 3. The two existing methods

In this section we will discuss the two existing methods of collocation extraction which are the basis for the proposed model.

*3.1. Mutual Information*

Mutual information (MI) from information theory has been utilized to find the closeness between word pairs by Church and Hanks [3]. Mutual information for two events x and y is defined as:

$$I(x, y) = \log_2 \frac{P(x, y)}{P(x).P(y)} \tag{1}$$

If we write $w_1$ and $w_2$ for the first and second word respectively, instead of x and y, then the mutual information for the two words $w_1$ and $w_2$ is given by:

$$I(w_1, w_2) = \log_2 \frac{P(w_1, w_2)}{P(w_1).P(w_2)} \tag{2}$$

where $P(w_1, w_2)$ is the probability of two words $w_1$ and $w_2$ coming together in a certain text and $P(w_1)$ and $P(w_2)$ are the probabilities of $w_1$ and $w_2$ appearing separately in the text, respectively.

If $P(w_1, w_2) = P(w_1).P(w_2)$, that is, the two words are independent to each other, then $I(w_1, w_2) = 0$, which indicates that these two words are not good candidates for collocation. A high mutual information score signifies the presence of a collocation.

*3.2. The t-test*

The t-test has been used by Church & Hanks [3] for collocation discovery to test the validity of a hypothesis. For that purpose, we formulate a null hypothesis $H_0$ that the two words $w_1$ and $w_2$ appear independently in the text. So under the null hypothesis $H_0$, the probability that the words $w_1$ and $w_2$ are coming together is simply given by:
$$P(w_1, w_2) = P(w_1).P(w_2).$$

The null hypothesis has been tested by using t-test. If the null hypothesis is accepted, we conclude that the occurrence of two words is independent of each other. Otherwise, we may conclude that they depend on each other, that is, they form collocations. In t-test we use the null hypothesis that the sample is drawn from a distribution with mean $\mu$, taking sample mean and variance into account. The t-test considers the difference between the observed and expected mean. The t statistic is defined as:

$$t = \frac{\overline{x} - \mu}{\sqrt{s^2/N}} \sim t_{n-1}(\alpha) \qquad (3)$$

where $\overline{x}$ is the sample mean, $s^2$ is the sample variance, $N$ is the sample size, $\mu$ is the mean of the distribution and $t_{n-1}(\alpha)$ denotes a t- distribution with (n-1) degrees of freedom at $\alpha$ level of significance. To apply t- test for testing the independence of two words $w_1$ and $w_2$, we assume that $f(w_1)$, $f(w_2)$ and $f(w_1, w_2)$ are the respective frequencies of the word $w_1$, $w_2$ and $w_1 w_2$ in the corpus and $N$ is the total number of words / bigrams in the corpus. Then, we have,

$$P(w_1) = \frac{f(w_1)}{N} \text{ (say } p_1\text{)}, \qquad P(w_2) = \frac{f(w_2)}{N} \text{ (say } p_2\text{)},$$

$$P(w_1, w_2) = \frac{f(w_1, w_2)}{N} \text{ (say } p_{12}\text{)},$$

The null hypothesis is

$$H_0: P(w_1, w_2) = P(w_1).P(w_2) = p_1.p_2$$

If we select bigrams (word pairs) randomly then the process of randomly generating bigrams of words and assigning 1 to the outcome that the particular word combination for which we are looking for is a collocation and 0 to any other outcome follows a Bernoulli distribution. For the Bernoulli distribution we have Mean $(\mu) = p$ and Variance $(\sigma^2) = p(1-p)$.

Thus, if the null hypothesis is true, the mean of the distribution is $\mu = p_1.p_2$. Also, for the sample, we have $P(w_1, w_2) = p_{12}$. Therefore, using Binomial distribution, sample mean $\bar{x} = p_{12}$ and sample variance $s^2 = p_{12}(1-p_{12})$. Using (3), we calculate the value of $|t|$ and compare it with the tabulated value at given level of significance. If the value of $|t|$ for a particular bigram is greater than the value obtained from the table, we reject the null hypothesis, which indicates that the bigram may be considered as a collocation. For the level of significance $\alpha = .005$, $|t| = 2.57$.

## 4. Fuzzy Inference System

Matlab Fuzzy Logic Toolbox provides an opportunity to look at all the components of FIS. It allows modifications, examination and verification of the effects of changes.

### 4.1. Baseline model

The collocation extraction Fuzzy inference system (CE-FIS) is based on the two existing techniques of collocation extraction, i.e., Mutual information and t-test which are the input variables for CE-FIS. Grade of membership is the out put variable.

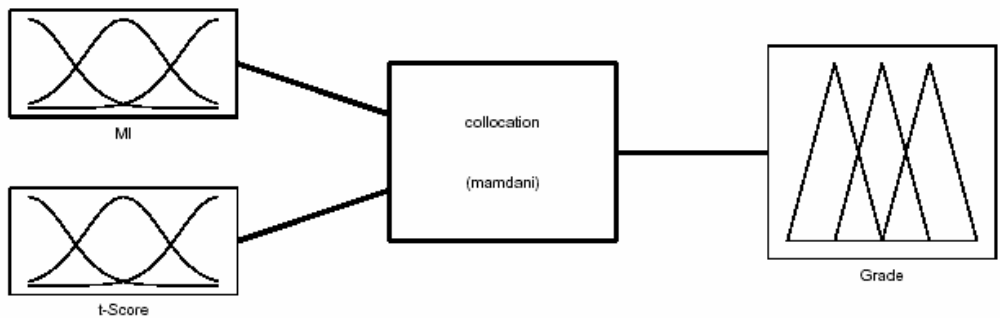

Fig1. Fuzzy inference system for collocation extraction

*4.2. Fuzzy sets / Membership Function*

Every input variable can be defined using two or three fuzzy sets. A membership function gives mathematical meaning to the linguistic variable such as high mutual information, low mutual information. Mutual information cab be defined using three fuzzy sets associated with each linguistic variable: high, average and low and t-score can be defined using two fuzzy sets associated with each variable: significant and non-significant. Output variable (relevance of a word combination for being collocation) can also be defined through three fuzzy sets namely: high, average and low. A membership function is a curve that defines how each point in the input/output space is mapped to degree of membership of fuzzy set. There are various inbuilt membership functions in fuzzy logic tool box. We have taken Pi shaped built-in membership function with different parameters for each input and output variable. Syntax of Pi shaped built-in membership is:

$$y = \text{pimf}(x, [a\ b\ c\ d])$$

This spline-based curve is so named because of its Π shape. This membership function is evaluated at the points determined by the vector x. The parameters a and d locate the "feet" of the curve, while b and c locate its "shoulders." Table 1 shows the chosen values of parameters 'a', 'b', 'c', 'd' for different membership functions.

Table 1: Values of parameters for different membership functions

| Membership function | value pf parameters | | | |
|---|---|---|---|---|
| | a | b | c | d |
| Low mutual information | -5.4 | -0.6 | 3 | 5.889 |
| Average mutual information | 3.002 | 5.922 | 6.175 | 9.092 |
| High mutual information | 6.143 | 8.94 | 13.1 | 17.9 |
| Non-significant t-score | -10.9 | -1.26 | -0.0159 | 2.778 |
| Significant t-score | -0.0476 | 2.68 | 14 | 23.6 |
| Low relevance | -0.45 | -0.05 | 0.184 | 0.5013 |
| Average relevance | 0.192 | 0.491 | 0.515 | 0.7976 |
| High relevance | 0.504 | 0.803 | 1.05 | 1.45 |

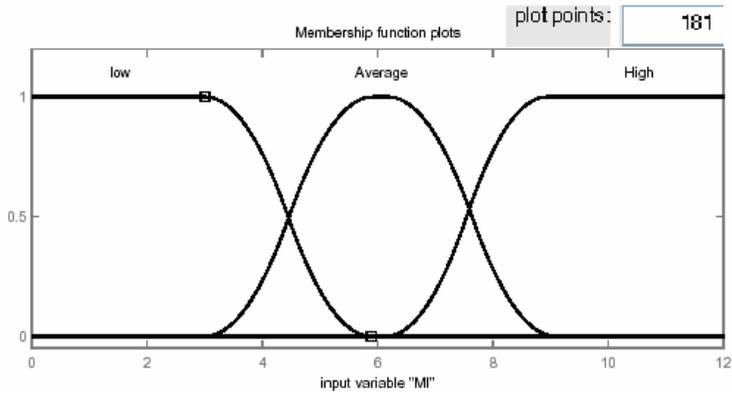
Fig.2. Membership functions for Mutual Information scores

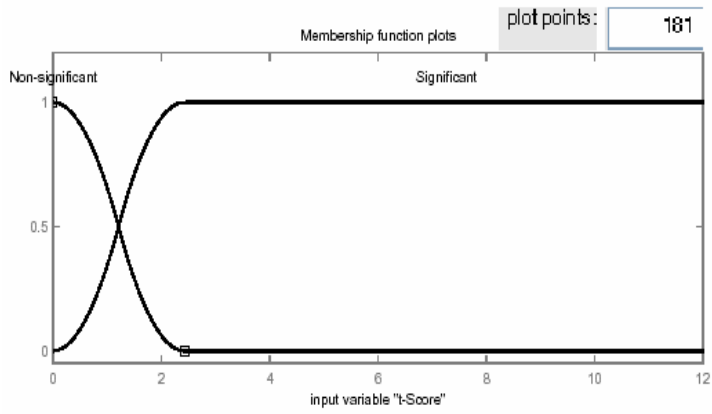
Fig.3. Membership functions for t-scores

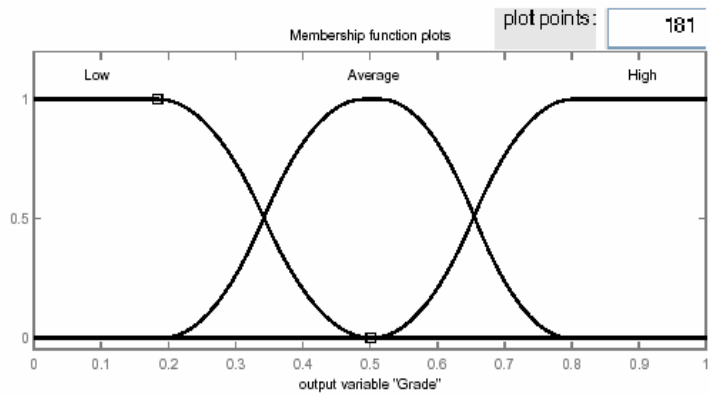
Fig.4. Membership functions for relevance

*4.3. Rules*

Matlab fuzzy toolbox allows defining rules by taking different fuzzy sets of input and output variables. Rules can be derived by simple reasoning of mutual information and t-score. High mutual information shows presence of a collocation and for a particular significance level, a word combination for which t-score is significant, can be considered for collocation. Following rules have been adopted for the CE-FIS.

- If (MI is low) and (t-score is non-significant), then (Relevance is low)
- If (MI is high) and (t-score is significant), then (Relevance is high)
- If (MI is low) and (t-score is significant), then (Relevance is average)
- If (MI is high) and (t-score is non-significant), then (Relevance is average)

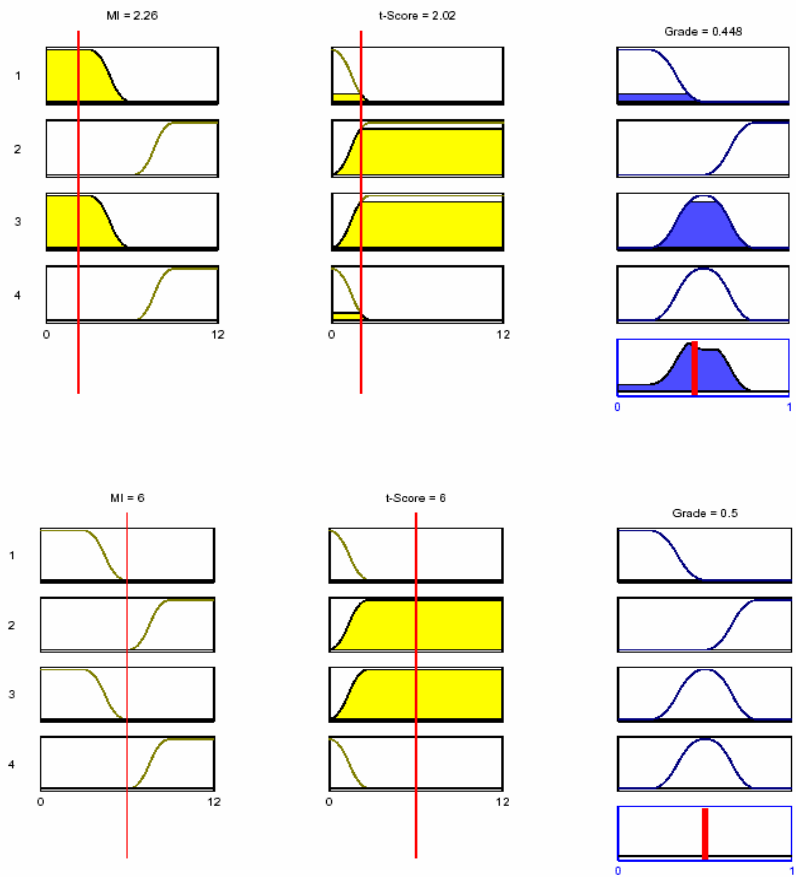

Fig. 5. Fuzzy Inference process

## 5. Evaluation

For evaluating the CE-FIS, we have compiled a corpus of one million words by taking some of the novels contained in project Gutenberg available at www.gutenberg.org/etext/<no.>(See appendix). To show the utilization of FIS, seventy word pairs from the compiled corpus have been examined for collocation as an example. Mutual information scores, t-scores and grades of membership using CE-FIS for each word combination have been calculated. Table 2 shows the frequencies $f(w_1)$, $f(w_2)$, $f(w_1w_2)$ of words $w_1, w_2$ and their combination $w_1w_2$ respectively, with their respective mutual information score, t-score. Table 3 shows the grades of membership of word combinations using CE-FIS.

Table 2: Mutual Information, t-score for different word combinations.

| $W_1$ | $W_2$ | $f(w_1)$ | $f(w_2)$ | $f(w_1w_2)$ | MI | $|t|$ |
|---|---|---|---|---|---|---|
| stark | madness | 6 | 22 | 1 | 12.96 | 1.00* |
| christmas | eve | 72 | 33 | 9 | 11.96 | 3 |
| base | camp | 54 | 55 | 7 | 11.27 | 2.64 |
| painful | experience | 27 | 87 | 3 | 10.39 | 1.73* |
| fire | bucket | 291 | 15 | 5 | 10.23 | 2.23* |
| valid | reason | 6 | 151 | 1 | 10.18 | 1.00* |
| spiritual | creature | 15 | 62 | 1 | 10.14 | 1.00* |
| rapid | motion | 32 | 42 | 1 | 9.61 | 1.00* |
| clumsy | fashion | 11 | 126 | 1 | 9.57 | 1.00* |
| public | opinion | 190 | 103 | 10 | 9.07 | 3.16 |
| visible | effort | 27 | 73 | 1 | 9.06 | 1.00* |
| empty | tent | 128 | 16 | 1 | 9 | 1.00* |
| huge | space | 35 | 64 | 1 | 8.87 | 1.00* |
| welcome | relief | 70 | 71 | 2 | 8.72 | 1.41* |
| peasant | girl | 10 | 253 | 1 | 8.7 | 1.00* |
| both | sides | 409 | 69 | 11 | 8.68 | 3.31 |
| national | guard | 39 | 163 | 2 | 8.37 | 1.41* |
| human | being | 182 | 735 | 30 | 7.88 | 5.45 |
| great | deal | 911 | 118 | 20 | 7.61 | 4.45 |
| wild | dreams | 136 | 46 | 1 | 7.39 | 0.99* |
| human | nature | 182 | 251 | 7 | 7.33 | 2.63 |
| little | episode | 1630 | 16 | 4 | 7.33 | 1.99* |
| evil | eye | 124 | 259 | 4 | 7.03 | 1.98* |
| dark | shadow | 279 | 93 | 3 | 6.92 | 1.72* |
| last | century | 846 | 32 | 3 | 6.86 | 1.72* |
| take | care | 808 | 228 | 20 | 6.83 | 4.43 |
| young | man | 741 | 2138 | 147 | 6.61 | 12 |
| cheerful | noise | 48 | 720 | 3 | 6.51 | 1.71* |
| early | days | 182 | 497 | 7 | 6.35 | 2.61 |
| last | link | 864 | 15 | 1 | 6.34 | 0.99* |
| long | journey | 967 | 100 | 7 | 6.25 | 2.61 |

| $W_1$ | $W_2$ | $f(w_1)$ | $f(w_2)$ | $f(w_1 w_2)$ | MI | $|t|$ |
|---|---|---|---|---|---|---|
| last | night | 846 | 856 | 47 | 6.09 | 6.76 |
| human | affairs | 182 | 86 | 1 | 6.07 | 0.99* |
| horrible | thing | 55 | 580 | 2 | 6.04 | 1.39* |
| trench | life | 99 | 1102 | 6 | 5.85 | 2.41* |
| water | level | 286 | 65 | 1 | 5.82 | 0.98* |
| usual | hour | 115 | 344 | 2 | 5.73 | 1.39* |
| step | towards | 135 | 348 | 2 | 5.48 | 1.38* |
| most | powerful | 723 | 33 | 1 | 5.46 | 0.98* |
| away | from | 862 | 3945 | 135 | 5.38 | 11.34 |
| good | terms | 1299 | 88 | 3 | 4.79 | 1.67* |
| came | along | 1360 | 367 | 13 | 4.77 | 3.47 |
| weary | night | 46 | 856 | 1 | 4.74 | 0.96* |
| great | emotions | 911 | 45 | 1 | 4.68 | 0.96* |
| make | use | 963 | 222 | 5 | 4.62 | 2.15* |
| every | night | 676 | 856 | 13 | 4.56 | 3.45 |
| strong | man | 172 | 2183 | 7 | 4.29 | 2.51* |
| night | before | 856 | 1164 | 18 | 4.25 | 4.02 |
| must | take | 1144 | 808 | 16 | 4.18 | 3.78 |
| almost | every | 518 | 676 | 6 | 4.17 | 2.31* |
| last | time | 846 | 1463 | 20 | 4.09 | 4.21 |
| except | myself | 81 | 1602 | 2 | 4.02 | 1.33* |
| wrong | way | 121 | 1084 | 2 | 4 | 1.33* |
| long | after | 967 | 1304 | 18 | 3.91 | 3.96 |
| night | air | 856 | 475 | 5 | 3.69 | 2.06* |
| only | because | 1187 | 371 | 5 | 3.58 | 2.05* |
| your | book | 2888 | 153 | 5 | 3.57 | 2.05* |
| another | half | 693 | 696 | 5 | 3.45 | 2.03* |
| come | over | 1358 | 1394 | 19 | 3.4 | 3.95 |
| little | chap | 1630 | 132 | 2 | 3.29 | 1.27* |
| look | upon | 756 | 1913 | 12 | 3.12 | 3.07 |
| things | behind | 622 | 415 | 2 | 3.03 | 1.24* |
| long | way | 967 | 1084 | 8 | 3 | 2.48* |
| looking | through | 431 | 1010 | 3 | 2.86 | 1.49* |
| like | myself | 1602 | 372 | 4 | 2.82 | 1.72* |
| along | over | 367 | 1394 | 3 | 2.62 | 1.45* |
| might | still | 1143 | 799 | 5 | 2.52 | 1.85* |
| round | upon | 387 | 1913 | 4 | 2.51 | 1.65* |
| make | sense | 963 | 192 | 1 | 2.51 | 0.82* |
| very | dark | 1410 | 279 | 2 | 2.42 | 1.15* |
| away | down | 862 | 1517 | 6 | 2.27 | 1.94* |

\* Rejected for collocation ( $|t| < 2.57$ )

The help of www.thefreedictionary.com and "Cambridge Advanced Learner's Dictionary" have been taken to check the validity of the word pairs and only 16 word pairs (star marked in table 3 & 4) have been found meaningful as collocations. On the basis of this we can compare the results of mutual information and t-score with the results obtained by model I and Model II. The mutual information does not provide a criterion for collocation extraction except saying high mutual information score shows the presence of a collocation. Precision and recall will depend upon the choice of the high mutual information score. However we can take different criteria of mutual information for calculating precision and recall. Table 5 shows the precision and recall for different mutual information scores. For t-score precision is 52% and recall is 69%. Table 6 and 7 show the precision and recall of the proposed models.

Table 3: Grades of membership for different word combinations using CE-FIS

| $w_1$ | $w_2$ | $D$ | $w_1$ | $w_2$ | $D$ |
| --- | --- | --- | --- | --- | --- |
| *christmas | eve | 0.82 | come | over | 0.50 |
| *public | opinion | 0.82 | long | after | 0.50 |
| *both | sides | 0.82 | last | time | 0.50 |
| *base | camp | 0.82 | must | take | 0.50 |
| *human | being | 0.81 | night | before | 0.50 |
| *great | deal | 0.80 | every | night | 0.50 |
| *human | nature | 0.79 | came | along | 0.50 |
| *fire | bucket | 0.79 | away | from | 0.50 |
| *take | care | 0.78 | long | way | 0.49 |
| *young | man | 0.77 | strong | man | 0.49 |
| painful | experience | 0.71 | almost | every | 0.47 |
| little | episode | 0.69 | night | air | 0.45 |
| welcome | relief | 0.66 | only | because | 0.45 |
| *national | guard | 0.66 | your | book | 0.45 |
| *evil | eye | 0.65 | another | half | 0.45 |
| *early | days | 0.64 | away | down | 0.44 |
| dark | shadow | 0.63 | make | use | 0.44 |
| last | century | 0.63 | like | myself | 0.41 |
| cheerful | noise | 0.62 | round | upon | 0.40 |
| wild | dreams | 0.62 | trench | life | 0.38 |
| last | link | 0.62 | *water | level | 0.38 |
| long | journey | 0.62 | usual | hour | 0.38 |
| stark | madness | 0.59 | step | towards | 0.37 |
| valid | reason | 0.59 | most | powerful | 0.37 |
| spiritual | creature | 0.59 | looking | through | 0.37 |
| rapid | motion | 0.59 | along | over | 0.37 |
| clumsy | fashion | 0.59 | good | terms | 0.36 |
| visible | effort | 0.59 | weary | night | 0.35 |
| empty | tent | 0.59 | except | myself | 0.34 |
| huge | space | 0.59 | wrong | way | 0.34 |
| peasant | girl | 0.59 | great | emotions | 0.34 |
| last | night | 0.50 | little | chap | 0.33 |
| horrible | thing | 0.50 | things | behind | 0.33 |
| human | affairs | 0.50 | very | dark | 0.31 |
| *look | upon | 0.50 | *make | sense | 0.26 |

*actual collocation

Table 4 : Precision and recall for Mutual Information.

| Mutual Information<br>( equal to or more than) | Precision<br>(in %) | Recall<br>( in %) |
|---|---|---|
| 10.0 | 43 | 19 |
| 8.0 | 35 | 38 |
| 6.0 | 38 | 81 |

Table 5 : Precision and recall for CE-FIS

| Grade of membership<br>( equal to or more than) | Precision<br>(in %) | Recall<br>( in %) |
|---|---|---|
| 0.80 | 100 | 38 |
| **0.70** | **91** | **63** |
| 0.60 | 59 | 81 |

## 6. Discussion

The present work was carried out to utilize the fuzzy inference system for collocation extraction. The previous two techniques were deterministic crisp formulas and it is difficult to make a decision about something which is vague and uncertain with deterministic crisp formulas. Fuzzy logic is based on the theory of fuzzy set which includes the elements with a grade of membership. Fuzzy logic for collocation extraction provides the benefits of the previous two approaches while overcoming their drawbacks. Observing table 2 and 3, some word combinations have high mutual information score (e.g., Stark madness) but have low grade of membership or rejected by t-test (e.g., fire bucket) but have high grade of membership. Table 5 shows the precision and recall of CE-FIS and it is very high in comparison to the two tests. Word combinations falling in the category of grade of membership more than 70 show high relevance to the set of collocations.

# Appendix

Following novels have been taken for corpus from Project Gutenberg.

1. Title: Bullets & Billets   Author: Bruce Bairnsfather [eBook #11232]
2. Title: Radio Boys Cronies,   Author: Wayne Whipple and S. F. Aaron [eBook  #11861]
3. Title: An Old Maid   Author: Honore de Balzac [eBook #1352]
4. Title: Behind the Line   Author: Ralph Henry Barbour [eBook #13556]
5. Title: The Light in the Clearing   Author: Irving Bacheller [eBook #14150]
6. Title: When William Came   Author: Saki [eBook #14540]
7. Title: The Marriage Contract   Author: Honore de Balzac [eBook #1556]
8. Title: The Historical Nights Entertainment, Second Series [eBook #7949]
9. Title: The Highwayman, Author: H.C. Bailey [eBook #9749]
10. Title: The Happy Foreigner  Author: Enid Bagnold  [eBook #9978]
11. Title: Mr. Bonaparte of Corsica    Author: John Kendrick Bangs  [eText   #3236]
12. Title: Love-at-Arms  Author: Raphael Sabatini  [eText #3530]
13. Title: The Toys of Peace    by H.H. Munro ("Saki") [eText #1477]
14. Title:  The Lion's Skin     Author:  Rafael Sabatini [eText #2702]
15. Title: The Lost City    by Joseph E. Badger, Jr. [eText #783]
16. Title:  The Muse of the Department  by Honore de Balzac [eText #1912]
17. Title: The Master of Silence Author: Irving Bacheller [eBook #7486]
18. Title: The Unbearable Bassington Author:  Saki [eBook #555]

*Note: Footnotes and illustrations have been removed while making corpus.*